# SEMANTIC FEATURE ANALYSIS: IMPROVING AGENTS WITHOUT SEARCHING OVER ROLLOUTS

**Yuval David**
IBM Software Innovation Lab
Haifa, Israel
yuval.david@ibm.com

**Fabiana Fournier**
IBM Software Innovation Lab
Haifa, Israel
fabiana@il.ibm.com

**Lior Limonad**
IBM Software Innovation Lab
Haifa, Israel
liorli@il.ibm.com

**Hadar Mulian**
IBM Software Innovation Lab
Haifa, Israel
Hadar.Mulian@ibm.com

## ABSTRACT

Ambiguity is an inherent property of natural-language agent specifications. When a system prompt leaves behaviour underdetermined, identical inputs follow divergent execution paths and produce inconsistent outcomes. The standard remedy is prompt optimisation: propose candidate prompts, run the agent to score them, and keep the best. This loop pays for the agent twice: once to generate candidates and again to rank them. On a tool-using agent whose rollouts cost dollars and minutes, the ranking cost dominates and budget-constrained optimisers routinely fail to find improvements.

We present Semantic Feature Analysis (SFA), a pipeline that repairs agent specifications *without running any search*. SFA reads execution traces the agent has already produced, clusters the outputs of each workflow node, decomposes them into semantic feature classes using an extended subject–verb–object schema, ranks those features by their contribution to outcome separation using a decision tree, and injects the surviving features as corrective statements into the affected node's system prompt. Because it never ranks candidate prompts, it never spends a rollout on selection.

We evaluate SFA against five prompt optimisers (GEPA, MIPROv2, SIMBA, BootstrapFewShot with random search, InferRules) and a single-reflection control, budget-matched in dollars at three budget levels across four benchmarks (IF-Bench, HotpotQA, HoVer, and GAIA). SFA consistently improves over the unmodified agent across benchmarks and budget levels, with the largest gains where rollouts are most expensive. On GAIA, where budget-constrained optimisers cannot afford to score even one candidate, SFA improves accuracy while other arms return their seed unchanged.

## 1 INTRODUCTION

An Agentic application is a software system that leverages large language models (LLMs) to act as autonomous agents, capable of perceiving their environment, reasoning through complex, multi-step problems, and taking action to achieve goals with minimal human intervention (Abou Ali et al., 2025). An agent's behaviour is governed by a *frame* (Dumas et al., 2023): a formulation of its goals and constraints, nowadays articulated in system prompts. Because those prompts are interpreted rather than executed, ambiguity in them may cause the same user input to follow different trajectories to different outcomes (Fournier et al., 2025). By an agent *trajectory* we mean the execution path from user input to final answer, including internally articulated system prompts, actions such as tool calls, and intermediate results.

The established way to repair a specification is to treat the prompt as an object to be optimized: propose candidates, score them by running the agent, and keep the best. This is the shape of automatic prompt engineering (Zhou et al., 2023; Yang et al., 2024) and of the optimizer family built around declarative language model programs (Khattab et al., 2024), including MIPROv2 (Opsahl-Ong et al., 2024), and of reflective evolutionary search (Agrawal et al., 2025).

Such a loop incurs two rollout costs. Each agent execution (a *rollout*) produces one scored candidate: one cost to generate candidates, and another to score them. The second cost is difficult to reduce without sacrificing performance. Using a cheaper selection mechanism increases the likelihood of selecting an inferior candidate. On the benchmarks where these methods are typically evaluated, this additional cost is easy to overlook: a rollout costs only about $0.009. For a tool-using research agent, however, each rollout is a multi-minute execution costing $0.22 on average (§5.3). At this scale, the cost of discriminating among candidates is no longer negligible and becomes a major factor in the overall cost and performance of the approach.

**Contribution.** We present the **Semantic Feature Analysis** (SFA) pipeline, which repairs the specification from traces the agent has *already* produced and never ranks candidate prompts. SFA mines multi-run execution logs with the aim of inferring corrective recommendations which are then injected into the system prompt. We elaborate the pipeline steps in section 3.

Prompt ambiguity is typically treated as a defect to be removed by better prompt writing. SFA reframes it as a resource. The same ambiguity that causes divergent trajectories also encodes, in the *semantic differences* between those trajectories, a precise account of what the prompt failed to say. It is that signal, rather than a search over candidate rewrites, that SFA harvests to derive corrective prompt injections. A second contribution is the rigorous comparative evaluation of this approach: we place SFA against five state-of-the-art prompt optimisers (GEPA, MIPROv2, SIMBA, BootstrapFewShot with random search, InferRules) and a single-reflection control, budget-matched in dollars across four benchmarks and three budget levels. The result is measurable and consistent improvement in agent performance (see §5): SFA significantly outperforms the unmodified program, a single reflection call, SIMBA and InferRules, and reaches parity with GEPA, MIPROv2 and BootstrapRS, all without spending a single additional rollout on candidate selection. The advantage is largest where it matters most: on the expensive-rollout GAIA benchmark, where budget-constrained optimisers cannot afford to score even one candidate, SFA improves accuracy by +5.2pp.

## 2 Related Work

**Prompts as specifications and the variability they generate.** In modern agent architectures, the LLM acts as a central controller that decomposes tasks, selects actions, and coordinates external tools (Xi et al., 2023; Wang et al., 2023; Karpas et al., 2022; Yao et al., 2023b; Schick et al., 2023). Prompt engineering surveys treat such prompts as executable specifications that require systematic design, testing, and maintenance (Schulhoff et al., 2024). Because decoding is stochastic and specifications are often incomplete, identical inputs can follow different execution paths and reach different outcomes (Tian et al., 2025; Winston & Just, 2025). Prior work responds to this instability either by sampling multiple outputs and selecting the best (Wang et al., 2022; Yao et al., 2023a) or by alignment tuning and constrained decoding (Ouyang et al., 2022; Bai et al., 2022; Beurer-Kellner et al., 2024). Both improve outcomes without touching the specification that causes the instability. SFA takes the opposite approach: it injects corrective statements directly into the system prompt of each affected node, aiming to resolve the ambiguity at its source.

**Observability, Diagnosis, and Trace Analysis** As agent pipelines grow more complex, observability has become essential for reliable deployment and for meeting service-level agreements (SLAs). Structured logging frameworks instrument agents to capture intermediate prompts, tool calls, and contextual states, enabling systematic monitoring and debugging AlSayyad et al. (2026). Complementary work on cognitive observability and failure taxonomies emphasizes reconstructing internal reasoning processes and categorizing root causes of breakdowns Rombaut et al. (2024); Winston & Just (2025); Moshkovich & Zeltyn (2025).

At the specification level, prompt defect analyses identify recurring sources of failure, including ambiguity, conflicting constraints, and missing requirements Tian et al. (2025). In parallel, multi-run analysis techniques draw on process mining and semantic clustering to align executions, localize deviations, and identify outcome-linked patterns van der Aalst (2016); de Leoni et al. (2015); Devlin et al. (2019); Reimers & Gurevych (2019); Lundberg & Lee (2017). Together, these approaches improve visibility and attribution in compound systems, but typically stop at diagnosis and do not generate reusable system-level corrections.

**Prompt optimization.** Automatic prompt engineering and optimization by prompting treat instruction design as search over natural-language programs (Zhou et al., 2023; Yang et al., 2024). Continuous methods instead optimise latent control vectors (Li & Liang, 2021). The DSPy family (Khattab et al., 2024) compiles declarative language model programs and provides several optimisers used here as baselines: MIPROv2 (Opsahl-Ong et al., 2024) searches instruction and demonstration pairs under a Bayesian trial budget, BootstrapFewShot[1] samples demonstration subsets, SIMBA[2] maintains a population of programs updated by rule- and demo-append steps, and InferRules (Tan et al., 2025) induces natural-language rules and retains the best by validation score. GEPA (Agrawal et al., 2025) evolves prompts through reflective mutation with a Pareto front over tasks. All of these run the agent to score candidates, while SFA derives corrective prompt injections from existing execution traces.

**Reflection.** Reflexion and Self-Refine store linguistic critiques or self-evaluations to guide later behaviour (Shinn et al., 2023; Madaan et al., 2023), but their adaptation is typically scoped to a single episode. In our experimental setup (§4), we include the cheapest reusable form of this idea as an explicit control arm: one reflection call over a sample of passing and failing runs, producing a rewritten prompt. It is the simplest intervention a practitioner would try, and it proves to be a competitive control (§5).

**Ambiguity repair and specification repair.** A related line of work targets specification ambiguity directly. ClarifyGPT (Mu et al., 2023) detects ambiguous requirements and asks clarifying questions before generating code. Work on underspecified software-engineering tasks shows large gains from such interaction (Vijayvargiya et al., 2025). SpecFix (Jia et al., 2025) repairs ambiguous problem descriptions automatically. Three differences separate this family from SFA. *Input*: they read a single task description. SFA reads a corpus of completed multi-node trajectories. *Object*: they rewrite the task given to the model. SFA rewrites the persistent system prompt of a node, which applies to every future task routed through it. *Loop*: ClarifyGPT and interactive methods query a human. SFA is fully automatic.

## 3 Semantic Feature Analysis

SFA is implemented as an analysis pipeline over an OpenTelemetry-based agent trace store. An anonymized implementation is included as a ZIP archive in the supplementary material. Its stages are those of Figure 1 and detailed below. A worked end-to-end example on real data (Figure 5) and the full illustrative walk-through on a small access-control agent are given in Appendix A.

### 3.1 Input and trajectory mining

The pipeline consumes the user prompt, the observed agent's specification (code and system prompts), and trajectory logs of ordered action executions with their payloads. Analysis uses the input and output *texts* of each action. Given $k$ runs we model the logs as process event logs of timestamped tasks and apply process mining (Fournier et al., 2025; van der Aalst, 2016) to elicit a *workflow view*, a graph structure in which nodes represent instances of the same action (i.e., the same trajectory input prompt and its output), and edges represent transitions between actions. Node identity is derived automatically by grouping LLM spans that share a system-prompt identity, so no per-agent configuration is needed to say which spans constitute a node.

### 3.2 Node clustering and trace selection

The classifier needs a binary run outcome, and SFA obtains it in one of two regimes.

[1]https://dspy.ai/api/optimizers/BootstrapFewShot/
[2]https://dspy.ai/api/optimizers/SIMBA/

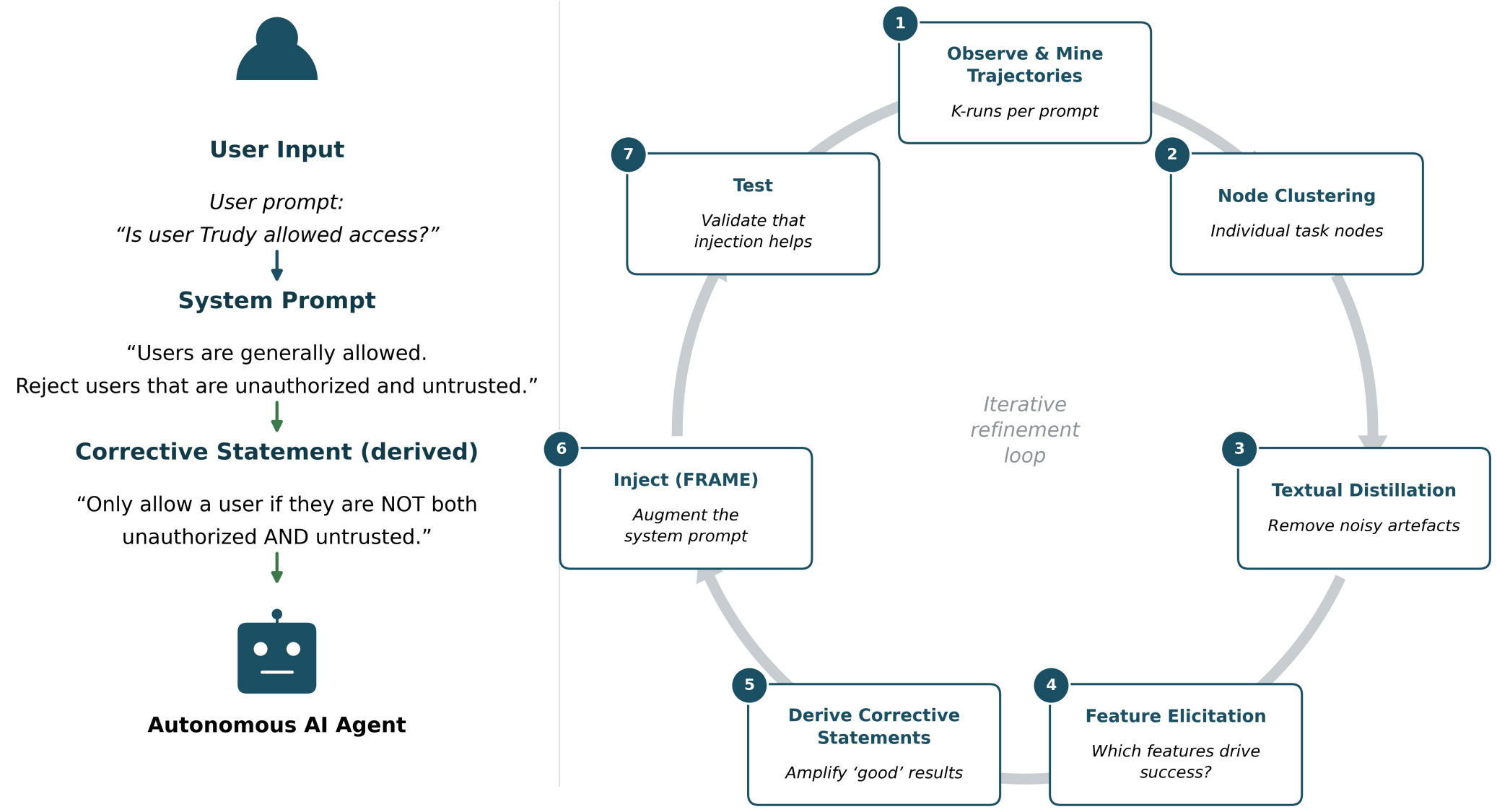

Figure 1: The improvement lifecycle. Multiple runs are mined into a workflow view; each node's outputs are clustered; text is distilled; semantic features are elicited and ranked by a decision tree; surviving features become corrective statements injected into that node's system prompt.

*Labelled, many tasks.* Where the agent already emits an outcome signal (a benchmark metric, a production success event, or a downstream acceptance), SFA consumes it directly across a corpus of many tasks. This is the regime of every experiment reported here: labels come from the benchmark's own pass/fail rating and *no human annotation is performed*.

*Unlabelled, one task repeated.* Where no oracle exists, the original setting for this pipeline, the corpus is instead many repetitions of a *single* task, and the signal comes from the behavioural spread across those repetitions. SFA embeds each node's output text, clusters it to separate the distinct behaviours the agent exhibits on that one task, names each cluster with an LLM, and asks once which is desirable (i.e., LLM-as-a-judge or an expert user). That single judgement then labels every run in the cluster.

### 3.3 Text distillation

In preparation for the semantic analysis of each workflow node's output text, we first preprocess the logged entries to remove noise introduced during log generation. This noise includes extraneous symbols and structural markers originating from the computational log representation.

We perform this preprocessing step via text distillation: an LLM is prompted to produce a clean, natural-language rendering of each entry that preserves its underlying semantics while discarding formatting noise.

Alternative approaches such as rule-based parsing or regular-expression filtering could remove such symbols, but they operate on surface form only. Leveraging an LLM instead enables semantic-aware normalisation, ensuring that the core meaning of the original text is retained during preprocessing.

### 3.4 Semantic feature elicitation

This step aims to elicit the core semantic components of each distilled sentence in order to explain the partitioning of node instances into the answer-node clusters identified above.

To this end, we rely on an extended SVO-based approach to decompose the grammatical structure of each sentence into meaningful components, referred to as *feature classes*. SVO (Subject–Verb–Object) denotes the most common fundamental sentence structure in English and many other languages: the agent performing the action (Subject) appears first, followed by the action itself (Verb), and finally the entity receiving the action (Object). We extend this schema to incorporate additional semantic clauses and elementary discourse units (EDUs), such as causes, conditions, and results.

For each node's output text, we sample 20 instances per cluster and instruct the LLM to identify the semantic feature classes that best distinguish instances across clusters (elicitation prompt in Appendix A). Once the feature classes are identified, we aggregate all textual instantiations of each class across all node instances, cluster them in the embedding space by semantic similarity, and assign descriptive labels to the resulting clusters with an LLM. This gives each feature class a discrete value range capturing the full range of expressions observed at that node. Each node instance is then encoded into a binary feature matrix: each row corresponds to one execution instance, each column to a possible feature value, and a cell is set to 1 if that value is present and 0 otherwise. Each row is additionally labelled with the answer-cluster assignment of its trace.

This matrix serves as input to a CART decision-tree classifier which uses Gini impurity (Loh, 2011), with the answer-cluster label as the target. The tree yields a ranking of feature values by their contribution to impurity reduction and its structure explains *which combinations* of features separate the clusters. Full implementation settings are in Appendix B.

After building the tree, the pipeline performs a quality check: it computes the tree's **cross-validated macro-F1** on held-out folds, weighting the failure class equally. If this score reaches cv-F1 $\geq$ 0.6, the pipeline trusts the tree's findings and proceeds to derive and inject corrective statements. If the score falls below this threshold, the extracted features do not reliably explain why some runs succeed and others fail, so the pipeline skips the injection step for that node entirely. This prevents changes from being made without sufficient evidential grounding.

### 3.5 Corrective statement derivation

In this step, we employ an LLM-as-a-judge prompt to derive corrective statements based on the semantic features identified in the previous step. These statements are formulated to mitigate ambiguities in the original system prompts and to improve performance with respect to the successful clusters identified during the answer-node clustering phase. Concretely, the LLM is prompted with (i) the system prompt of the node under examination, (ii) the elicited features with their descriptions and importance scores, (iii) the decision-tree rules, and (iv) an enumeration of the successful clusters, and asked to recommend prompt improvements that steer the agent toward the success patterns. The resulting statements are injected into that node's prompt. Nothing is re-scored: the pipeline emits one rule set and stops.

## 4 Experimental Setup

To evaluate the SFA pipeline, we conducted an empirical comparative study placing it against state-of-the-art prompt optimisation alternatives. Specifically, we compared SFA against five prompt optimisers (GEPA, MIPROv2, SIMBA, BootstrapFewShot with random search, and InferRules), a single-reflection control, and an unmodified baseline, across four benchmarks and three budget levels. All arms were given the same monetary budget and the same training tasks, and performance was measured on a held-out test set. This design allows a direct, controlled assessment of whether SFA's trace-grounded approach to specification repair is competitive with methods that actively search over candidate prompts.

The comparison was budget-matched in dollars: every arm received the same budget $B$, derived from SFA's own measured analysis cost, and was stopped when it reached that cap. All arms trained on the same tasks and were evaluated on the same held-out test set. Trace collection was not charged to $B$, as collecting execution traces is a prerequisite shared by all compared techniques in order to establish the performance baseline. Test evaluations were likewise charged to no arm.

The budget was varied across three levels by increasing the number of repetitions per training task, while keeping the task set and agent fixed, isolating the effect of budget.

Not all traces can be read within a fixed budget. Top-$k$ selection keeps only the most variable tasks and drops tasks where every run failed. Those all-fail tasks carry a uniform failure mode that contrastive analysis cannot detect. We therefore use *balanced sampling*, which retains all tasks regardless of their pass rate. A complementary node-level uniformity check flags defects present in every run, which contrastive analysis misses by design.

To ensure a fair comparison, the agent under study is never analysed by itself. The lightweight stages (distillation, clustering, and feature elicitation) run on claude-haiku-4-5, while corrective-statement generation and rule-set assembly run on claude-sonnet-4-6, the same model used by GEPA and the control arm for reflection. No arm is therefore advantaged by model choice. All stages up to and including rule-set assembly are charged to the budget $B$. The trailing human-readable summary is not.

Each benchmark in this study is a combination of a public task dataset and a purpose-built agent whose system prompts SFA is asked to improve. The task dataset defines what a correct answer looks like. The agent is the multi-node program that produces answers and whose prompts are modified. The study covered four such benchmarks:

- **IFBench** (Pyatkin et al., 2025): a public instruction-following dataset with verifiable constraints, paired with a 2-node DSPy agent on cheap rollouts.
- **HotpotQA** (Yang et al., 2018): a public multi-hop question-answering dataset, paired with a 4-node DSPy agent on cheap rollouts.
- **HoVer** (Jiang et al., 2020): a public three-hop claim-verification dataset (scored as document retrieval recall), paired with a 4-node DSPy agent on cheap rollouts.
- **GAIA** (Mialon et al., 2023) level 1: a public research-task dataset, paired with a ReAct agent equipped with live web search and a 50-step limit, representing the expensive-rollout regime.

Eight arms were compared: SFA, GEPA, MIPROv2, SIMBA, BootstrapFewShot with random search, InferRules, and control (a single LLM reflection call that reads a sample of 20 failing and 10 passing runs and rewrites the prompt once, without any search or iteration). All arms were evaluated against base, the unmodified agent with no prompt changes applied.

Budgets were enforced against a live dollar ledger rather than a fixed rollout count. This meant each run's actual API spend was accumulated in real time, and a run was stopped as soon as the dollar cap was reached, regardless of how many rollouts that consumed.

The whole study (every arm, every budget point, every pilot and superseded run) cost \$946.38 measured from the ledgers, of which SFA's analysis was \$123.69, baseline search \$307.79, and held-out evaluation, charged to no arm, \$514.90. Evaluation is not folded into $B$.

## 5 Results

### 5.1 Comparative scores across benchmarks

Table 1 reports held-out test scores for all eight arms at three budget levels across all four benchmarks. For the three cheap-rollout benchmarks (IFBench, HotpotQA, HoVer), budget levels $B_1$, $B_2$, and $B_3$ correspond to 1, 2, and 3 repetitions per training task respectively. For GAIA, the expensive-rollout benchmark, each reported score is the mean over a 210-run held-out evaluation, and $B$ covers SFA's analysis plus any additional traces beyond those already collected. Across all twelve budget-benchmark columns, SFA achieved the best average rank (1.92), outranking every other arm when performance is aggregated over benchmarks and budgets.

### 5.2 Statistical analysis

The dependent variable for every analysis was the per-task binary outcome (pass/fail), recorded for each arm on the held-out test set. Each reported score is the mean of these outcomes. For the three DSPy benchmarks this was 294–300 tasks evaluated once per arm. For GAIA it was 42 tasks × 5 repetitions = 210 outcomes per arm.

The independent variables were *arm* (eight arms, manipulated), *budget* (three levels, manipulated), and *benchmark* (four benchmarks, observed). Because every arm was evaluated on the same held-out tasks, each task produces one binary outcome per arm. Comparisons between arms were made within tasks: arm A is compared to arm B by examining how often A passes where B fails on the same task, and vice versa, rather than by comparing their overall means as independent samples. This removes task-difficulty variance from the error term and makes the comparisons more sensitive.

Table 1: Held-out test scores at three budget points on all four benchmarks. Budgets are all-in and are *per benchmark*; dollar figures are not comparable across benchmarks. The subscript after each score is that arm's rank within that column (1 = best; ties share a rank). Bold marks the highest score in each budget column and the lowest (best) mean rank. Avg. Rank is each arm's mean rank across all twelve columns. †search ran but returned the seed unchanged.

| | **IFBench** | | | **HotpotQA** | | | **HoVer** | | | **GAIA** | | | **Avg.** |
|---|---|---|---|---|---|---|---|---|---|---|---|---|---|
| **arm** | \$4.50 | \$5.85 | \$8.54 | \$8.87 | \$11.57 | \$14.27 | \$8.87 | \$14.27 | \$19.58 | \$6.98 | \$53.67 | \$96.60 | **Rank** |
| base | .4405$_{(4)}$ | .4405$_{(6)}$ | .4405$_{(6)}$ | .4100$_{(7)}$ | .4100$_{(7)}$ | .4100$_{(7)}$ | .4933$_{(4)}$ | .4933$_{(7)}$ | .4933$_{(7)}$ | .700$_{(3)}$ | .700$_{(7)}$ | .700$_{(8)}$ | 6.08 |
| **SFA** | .5357$_{(2)}$ | **.5952**$_{(1)}$ | **.5918**$_{(1)}$ | .5733$_{(5)}$ | .7033$_{(2)}$ | **.7333**$_{(1)}$ | .5367$_{(2)}$ | .5200$_{(3)}$ | .5233$_{(3)}$ | **.752**$_{(1)}$ | **.790**$_{(1)}$ | **.771**$_{(1)}$ | **1.92** |
| GEPA | .5119$_{(3)}$ | .5119$_{(2)}$ | .4881$_{(4)}$ | .6533$_{(3)}$ | **.7100**$_{(1)}$ | .7100$_{(4)}$ | .4633$_{(8)}$ | .5184$_{(5)}$ | .5167$_{(4)}$ | .700†$_{(3)}$ | .738$_{(3)}$ | .738$_{(2)}$ | 3.50 |
| MIPROv2 | .4405$_{(4)}$ | .4405$_{(6)}$ | .4405$_{(6)}$ | .6933$_{(2)}$ | .6933$_{(4)}$ | .6933$_{(6)}$ | **.5667**$_{(1)}$ | **.5667**$_{(1)}$ | **.5933**$_{(1)}$ | .700†$_{(3)}$ | .7286$_{(5)}$ | .7286$_{(3)}$ | 3.50 |
| SIMBA | .4405$_{(4)}$ | .4405$_{(6)}$ | .4405$_{(6)}$ | .4100$_{(7)}$ | .5800$_{(6)}$ | .7167$_{(2)}$ | .4933$_{(4)}$ | .5067$_{(6)}$ | .5067$_{(5)}$ | .700†$_{(3)}$ | .7429$_{(2)}$ | .7286$_{(3)}$ | 4.50 |
| InferRules | .4337$_{(8)}$ | .4592$_{(5)}$ | .4456$_{(5)}$ | .6533$_{(3)}$ | .6700$_{(5)}$ | .7167$_{(2)}$ | .4967$_{(3)}$ | .4867$_{(8)}$ | .4667$_{(8)}$ | .700†$_{(3)}$ | .7143$_{(6)}$ | .7143$_{(6)}$ | 5.17 |
| BootstrapRS | .4405$_{(4)}$ | .4966$_{(4)}$ | .4932$_{(3)}$ | **.7000**$_{(1)}$ | .7000$_{(3)}$ | .7000$_{(5)}$ | .4933$_{(4)}$ | .5567$_{(2)}$ | .4967$_{(6)}$ | .700†$_{(3)}$ | .7333$_{(4)}$ | .7095$_{(7)}$ | 3.83 |
| control | **.5360**$_{(1)}$ | .5102$_{(3)}$ | .5289$_{(2)}$ | .4867$_{(6)}$ | .2567$_{(8)}$ | .2100$_{(8)}$ | .4900$_{(7)}$ | .5200$_{(3)}$ | .5733$_{(2)}$ | .7095$_{(2)}$ | .6952$_{(8)}$ | .7238$_{(5)}$ | 4.58 |

Benchmark moderated the arm effect: which arm wins depends on which benchmark it runs on. Naive pooling across benchmarks would therefore understate the standard error, so all aggregate effects were estimated with a random-effects meta-analysis over the twelve (benchmark × budget) cells. Arms that exhausted their budget without producing a candidate returned the unmodified baseline program, so their per-task outcomes are identical to the baseline's.

An ANOVA over all 21,455 per-task outcomes on the three DSPy benchmarks, with task as a blocking factor, found a large method effect (i.e., arm; $F_7 = 97.5$, $p < .001$) and a smaller budget effect ($F_2 = 15.1$, $p < .001$). The ANOVA confirmed that the within-task paired design was necessary: task difficulty alone accounted for 64% of all variance, meaning unpaired comparisons would have been dominated by noise unrelated to the arms being compared. The dominant effect was the **method × benchmark interaction** ($F_{14} = 92.2$, $p < .001$, $\eta^2 = 0.020$), which exceeded the method main effect. A formal model comparison confirmed it was necessary ($F = 90.5$, $p < .001$). Which optimizer wins depends on the benchmark, not on any property of the optimizer alone.

Because the arm effect differed by benchmark, we conducted pairwise comparisons of each arm against SFA separately per benchmark and budget cell, then aggregated across cells using a random-effects meta-analysis (Table 2). This approach allows the true effect to differ across benchmarks rather than assuming a single pooled estimate.

Table 2: Random-effects aggregate over all four benchmarks and all budget points ($k$ cells). $I^2$ measures result consistency across benchmarks; values above 75% mean the effect size varied substantially across benchmarks. Bold $p$ values are significant ($p < .05$).

| **SFA vs** | $k$ | $\Delta$ | **95% CI** | $p$ | $I^2$ |
|---|---|---|---|---|---|
| base (unmodified) | 12 | +0.124 | [0.069, 0.179] | $\mathbf{9.8 \times 10^{-6}}$ | 93% |
| control (1 reflection call) | 12 | +0.115 | [0.026, 0.204] | **0.011** | 97% |
| SIMBA | 12 | +0.077 | [0.042, 0.112] | $\mathbf{1.4 \times 10^{-5}}$ | 83% |
| InferRules | 12 | +0.055 | [0.022, 0.089] | **0.0011** | 82% |
| GEPA | 12 | +0.031 | [0.008, 0.054] | **0.0089** | 74% |
| BootstrapRS | 12 | +0.033 | [−0.002, 0.069] | 0.064 | 85% |
| MIPROv2 | 12 | +0.028 | [−0.019, 0.075] | 0.241 | 92% |

SFA significantly outperformed the unmodified program, a single reflection call, SIMBA, GEPA, and InferRules across every benchmark and budget in the study. Against MIPROv2 and BootstrapRS the aggregate effect was positive but not statistically separable from zero. On GAIA, SFA led every baseline by 4.8–9.5pp and beat base significantly (+9.1pp, $p < .01$), though with 42 tasks no individual baseline contrast survived multiple-comparison correction. SFA does not score highest in every cell. The claim the evidence supports is that SFA reached parity-or-better with search-based optimizers *while spending no rollouts on candidate selection*, and it outperformed both the unmodified program and the single-reflection baseline a practitioner would otherwise reach for.

### 5.3 Observations

The following observations highlight the most noteworthy patterns across the four benchmarks.

**(1) SFA led IFBench at every budget** (+0.3/ + 8.3/ + 6.3pp over the best competing arm at each tier, with the middle margin statistically separable). MIPROv2 and SIMBA returned the unmodified program at all three budgets. An uncapped MIPROv2 run at $15.95 reached only 0.5017, below SFA at the lowest budget.

**(2) SFA climbed fastest on HotpotQA** (0.5733 → 0.7033 → 0.7333, +16.0pp) and led at the top budget by 1.7pp over SIMBA (not statistically separable). MIPROv2, BootstrapRS, and GEPA were partly or fully frozen across the budget curve.

**(3) SFA ranked fifth on HotpotQA at the lowest budget** (0.5733). At one repetition there is no same-task pass/fail contrast, the core signal SFA relies on, so arms that do not require it outperformed SFA at this tier.

**(4) SFA did not win HoVer.** MIPROv2 led at every budget (0.5933 vs. SFA's 0.5233 at the top tier). HoVer measures retrieval recall. Instructions change what the agent does with retrieved evidence, not which documents come back. BootstrapRS peaked at $B_2$ (0.5567) but fell back at $B_3$ (0.4967), showing that few-shot demonstrations do not consistently help on retrieval.

**(5) The single-reflection control was competitive but unstable.** `control` outperformed MIPROv2, SIMBA, and InferRules across all IFBench budgets, yet *collapsed* on HotpotQA from 0.4867 to 0.2100, far below the base of 0.4100. Its three cells estimate single-shot rewrite variance (HotpotQA 27.7pp, HoVer 8.3pp, IFBench 2.6pp).

**(6) On GAIA, SFA led at every budget.** At $6.98, five of seven arms returned their seed unchanged, while SFA delivered 0.752 against a base of 0.700. At higher budgets SFA continued to lead (0.790 at $53.67, 0.771 at $96.60), with the slight drop suggesting saturation. At $53.67, GEPA produced format rules only, while SFA additionally generated tool-usage workarounds, accounting for the 5.2pp gap.

### 5.4 Ablation study: contribution of each pipeline design choice

To assess which design choices drive accuracy, we ran a staged ablation on gpt-oss-120b. Each variant removes one element of the pipeline while keeping everything else fixed, then regenerates corrective statements from the same recorded runs and evaluates on the same held-out tasks (IFBench, HotpotQA, and HoVer: three draws over 300 tasks; GAIA: 42 tasks × 5 repetitions). Table 3 reports the contribution of each choice: the accuracy (i.e., mean gain) of the full pipeline minus the accuracy of the pipeline with that choice removed.

Table 3: What each design choice contributes on gpt-oss-120b: points of test score the reported SFA rule set gains over the same pipeline with that choice removed. One choice is removed per row.

| contribution of | IFBench | HotpotQA | HoVer | GAIA | mean |
|---|---|---|---|---|---|
| tree-based ranking | -2.8 | 0.0 | +3.8 | +8.6 | +2.4 |
| full-corpus contrast | -0.4 | +5.6 | -1.6 | +0.5 | +1.0 |
| semantic feature evidence | +2.3 | +7.8 | +1.4 | +0.5 | +3.0 |
| clause guards + node contracts | +0.2 | +3.4 | +10.9 | +2.4 | +4.2 |
| balanced trace selection | +0.9 | +4.3 | +7.3 | +2.4 | +3.7 |
| own-model corrective statements | +2.8 | +12.6 | -0.1 | +0.5 | +3.9 |
| ungated injection | +3.6 | +19.0 | +1.8 | -1.0 | +5.9 |

*Tree-based ranking.* The LLM-derived feature classes are passed to the corrective statement derivation step as usual, but the decision tree's outputs are withheld: rule paths, feature importances, and decisive signals are not provided. This measures what tree-based ranking adds over LLM feature elicitation alone.

*Full-corpus contrast.* Each feature's pass rate over all recorded runs is provided to the corrective statement derivation step alongside the tree ranking. This variant removes it, measuring its contribution over the tree signal alone.

*Semantic feature evidence.* All feature-derived inputs are withheld at once: values, descriptions, tree, and contrast. The corrective statement derivation step receives only the labelled node outputs and the node's current prompt. This variant isolates the effect of the feature elicitation step.

*Trace selection.* The reported system samples a fixed number of traces per task, balancing pass and fail examples within each task and retaining all tasks including all-fail and all-success ones. This variant replaces that with top-$k$ task selection, which ranks tasks by their pass/fail variance and keeps only the $k$ most variable.

*Corrective statements derivation model.* The corrective statement derivation model changes: distillation, feature elicitation, and the tree run on the agent's own model, but the same evidence is passed to Llama-3.1-405B for the writing step.

*Ungated injection.* This ablation compares two injection policies: injecting corrective rules from all trees regardless of their fit, versus injecting only from trees whose cross-validated F1 exceeds 0.6. The reported system is ungated (all trees), which proved the better policy.

## 6 Discussion and Conclusion

The central claim of this paper is that ambiguity in agent specifications is not merely a defect to be corrected by better prompt writing: it is a signal. The same divergence it induces in execution traces encodes, in the semantic differences between those trajectories, a precise account of what the prompt failed to say. SFA harvests that signal without running the agent a single additional time, and the results confirm it. Across four benchmarks and three budget levels, SFA either matches or outperforms every search-based baseline, and it does so with a qualitatively different cost profile: no inner loop, no candidate ranking, no compounding selection noise.

The advantage is not incidental to budget. When rollouts are cheap, search is viable and the difference narrows. When each rollout is an expensive, multi-minute tool-using execution, search-based methods exhaust their budgets before they can reliably distinguish a good candidate from a noisy one. SFA is immune to this failure mode by construction: having no selection loop, it has no loop in which noise can accumulate. The superiority of SFA on the expensive-rollout GAIA benchmark is therefore the expected consequence of the design.

The failure modes SFA repaired validate the effectiveness of its mechanism. The taxonomy recovered from the training corpora (Table 6, Appendix C) reveals that under-specification is the dominant failure class across all three benchmarks, accounting for the large majority of repaired traces. Under-specification is itself a form of ambiguity: a prompt that omits a verification step or under-constrains a multi-hop query is one whose intent remains partially undefined. Ambiguity, treated here as a resource rather than a defect, turns out also to be the dominant failure mode the pipeline repairs. SFA does not merely handle one narrow class of defect: it addresses the broad, pervasive category of prompts that leave agent behaviour underdetermined, which is precisely the category most resistant to repair by conventional prompt optimisation. The same property of agent specifications that motivated the pipeline is the one the pipeline demonstrably fixes.

## 7 Limitations and Future Work

**Feature derivation stability is unmeasured.** Re-evaluating a fixed rule set gave a ~2pp band on the benchmarks (five runs on GAIA, three on the others). What remains unmeasured is re-derivation: no SFA analysis has been re-run on a resampled corpus, so we cannot report whether the elicited feature classes, the induced tree, or the resulting rules are stable across draws. Future work should re-run the full analysis pipeline on resampled corpora to establish derivation variance alongside the output variance reported here.

**Prompt growth and cross-node conflict.** SFA appends rules to node prompts with no length budget or conflict check. Without a length cap, prompt token costs may inflate unboundedly as rules accumulate across iterations. Separately, one rule-set version contained directives in which one node's rule contradicted another's. Neither effect is measured systematically. Adding a prompt length budget and a rule-set lint step are straightforward improvements for future versions.

**Numeric values and rule redundancy.** The pipeline treats instructions as text and is agnostic to numeric values they may encode (thresholds, counts, weights). Such values cannot be represented as features in the decision-tree input, so the pipeline cannot surface or repair numeric underspecification. Separately, because the pipeline scans all nodes without reachability-causal awareness, it can emit rules for nodes that are downstream consequences of an upstream fault rather than root causes. This may produce redundant or misplaced rules. Both are directions for future work.

### Reproducibility Statement

An anonymized implementation is included as a ZIP archive in the supplementary material.

The archive contains the method, the benchmark harness (benchmark programs, datasets, metrics, and baseline optimizers), the shared evaluation loop that scores every arm the same way, and the scripts that reproduce each reported experiment. The archive also contains a README with steps to reproduce the reported results.

The pipeline code included in the ZIP archive is also released under an open-source license, with a link provided in the camera-ready version of this paper.

### Use of Large Language Models

LLMs were used to aid and polish writing and to draft sections of the paper. All experimental design, analysis, and conclusions are the authors' own.

## A Worked Example: the AccessControl Agent

This section illustrates the SFA pipeline stages on a concrete example small enough to trace end to end. The application manages access to protected resources. It consists of an orchestrator agent equipped with two tools, check_unauthorized_users and check_untrusted_users, each instantiating a sub-agent that verifies whether the user is authorised and trusted, respectively. The orchestrator's system prompt reads:

```
Users are generally allowed. Reject users that are
unauthorized and untrusted.
```

This prompt contains an inherent ambiguity: it is unclear whether the condition is conjunctive (reject only when both hold) or disjunctive (reject when either holds). The application was executed 100 times with a test user named Trudy, defined as authorised but untrusted, yielding 98 valid samples. Each run used the prompt ``Is user Trudy allowed access?''. The expected outcome is rejection.

### A.1 Preliminary Configuration

For the AccessControl agent example, we employed gpt-oss-120b as the backbone LLM, informed by a small-scale ablation study of model effectiveness shown in Table 4. We selected this model because of its favorable cost–performance trade-off, while the study provided evidence that the pipeline was largely agnostic to the choice of model family.

Table 4: Backbone sweep on AccessControl: pre- and post-injection accuracy for four backbone LLMs. The pipeline itself is unchanged across rows.

| **LLM** | **Pre** | **Post** |
|---|---|---|
| GPT-4o | 0.50 | 0.99 |
| LLaMA 4 Maverick | 0.50 | 0.89 |
| Mistral Medium | 0.50 | 0.86 |
| gpt-oss-120b | 0.50 | 0.87 |

### A.2 Trajectory Mining

Execution trajectories are modelled as process event logs of timestamped LLM invocations. Using process mining techniques, the logs are aggregated into a workflow view that reflects the task-flow structure across all runs. Each node in the view corresponds to a recurring action, and its instances capture the input and output texts of every execution of that action.

For the AccessControl example, mining yields a three-node workflow view (Figure 2): the orchestrator node, an XOR split over the two checker sub-agents, and a loop back to the orchestrator for the final access decision.

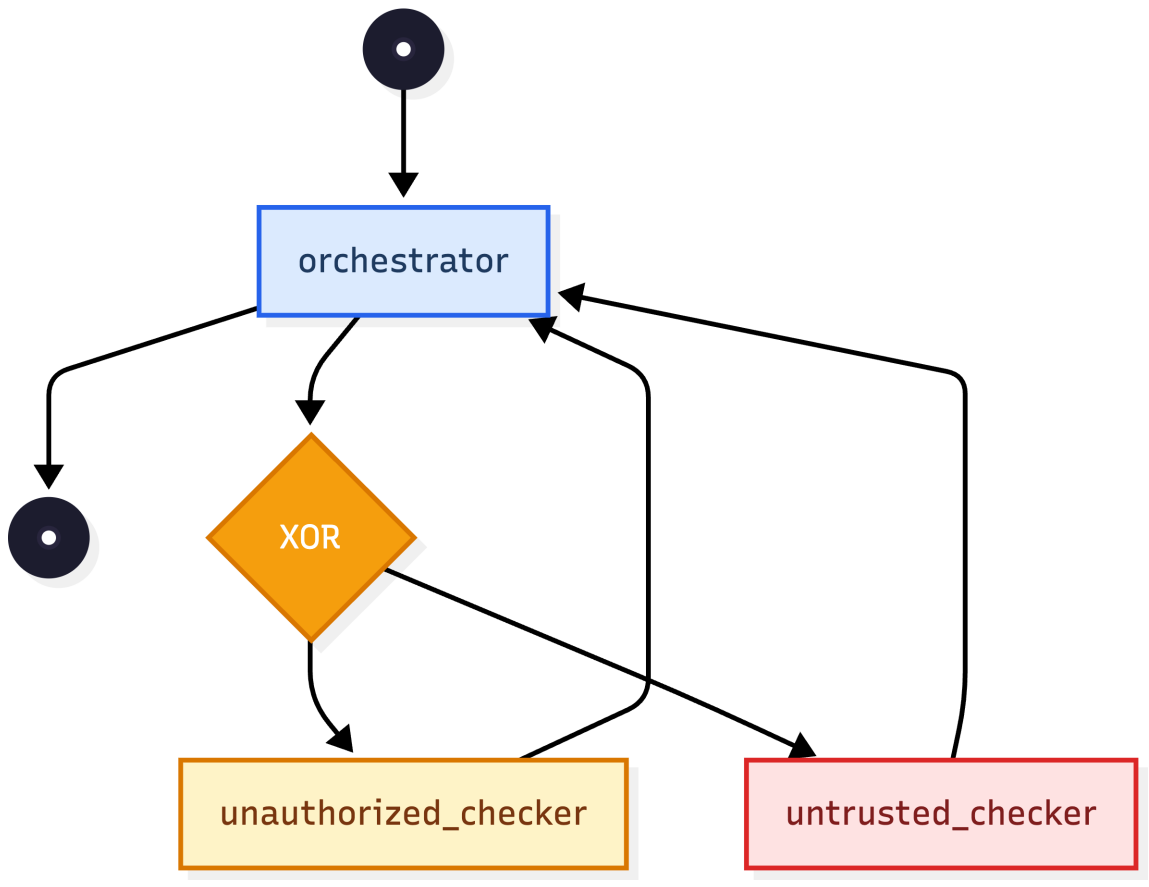

Figure 2: Workflow view mined from the AccessControl trajectories.

### A.3 Node Clustering

To distinguish passing from failing runs, the output texts of the answer node are embedded and partitioned using $k$-means clustering. The number of clusters is selected by an inertia-based elbow criterion. Each cluster is annotated with an LLM-generated representative label based on a sample of three traces.

In the AccessControl example, the answer node yields five clusters:

- “Allow unless both unauthorized and untrusted” (58 samples)
- “Apology and refusal” (37 samples)
- Three singletons (1 sample each)

As this example has no ground truth for its outcomes (i.e., unlabelled), the user identified cluster 1 as the desirable outcome. The three singletons were treated as negligible.

### A.4 Text Distillation

Before semantic analysis, each node’s output text is distilled by an LLM to remove formatting artefacts introduced by the logging layer, producing a clean natural-language sentence that preserves the original semantics.

```
"Summarize this log entry into one concise English sentence describing the action.\n"
"Focus on the VERB (action) and the OBJECT (target).\n"
f"LOG:\n{safe_text}"
```

Listing 1: Distillation prompt.

### A.5 Semantic Feature Elicitation

For each node, the pipeline samples 20 instances per cluster and instructs an LLM to identify semantic feature classes that distinguish instances across clusters, using an extended subject–verb–object (SVO) schema that also captures clauses, conditions, and causes. The LLM returns a JSON list of feature class definitions.

```
Your task is to elicit a list of semantic "features" that capture the core
meaning structure of these texts.

A feature here means a semantic class that can be instantiated by different
spans in different sentences, such as:
  Subject / "Who?"       Action / "What did they do?"
  Object  / "What was acted on?"      Reason / "Why?"

Each feature should correspond to a meaningful part of the sentence
(e.g., S-V-O roles, clauses, EDUs, causes, results, conditions).

Please focus on features that will differentiate between the following clusters:
<20 sampled traces per cluster>

Output as JSON containing the feature definitions.
Ensure descriptions emphasize text values and forbid numbers.
```

Listing 2: Feature-class elicitation prompt (abridged).

For the orchestrator node, the elicited feature classes include `refusal` mode, `condition expression`, `permission status`, `list status`, and others. The textual instantiations of each class are embedded and grouped into discrete values, producing the binary feature matrix of Figure 3: one row per run, one column per feature value, with a 1 if the value is present.

| | Refusal_mode_is_none | Condition_expression_is_none | ... | Invitation_content_is_ask_anything | action_is_allow | ... | | | Result Cluster |
|---|---|---|---|---|---|---|---|---|---|
| 0 | 1 | 0 | ... | 1 | 0 | | | | 0 |
| 1 | 0 | 1 | ... | 0 | 1 | | | | 1 |
| ... | | | | | | | | | ... |

Figure 3: Each trajectory becomes a binary vector of semantic feature values.

The matrix is the input to a CART decision tree trained to classify run outcomes. The tree ranks features by their Gini impurity reduction. Figure 4 shows the tree for the AccessControl example: the two most discriminative features are **refusal** mode (Gini 0.507) and **condition expression** (Gini 0.329).

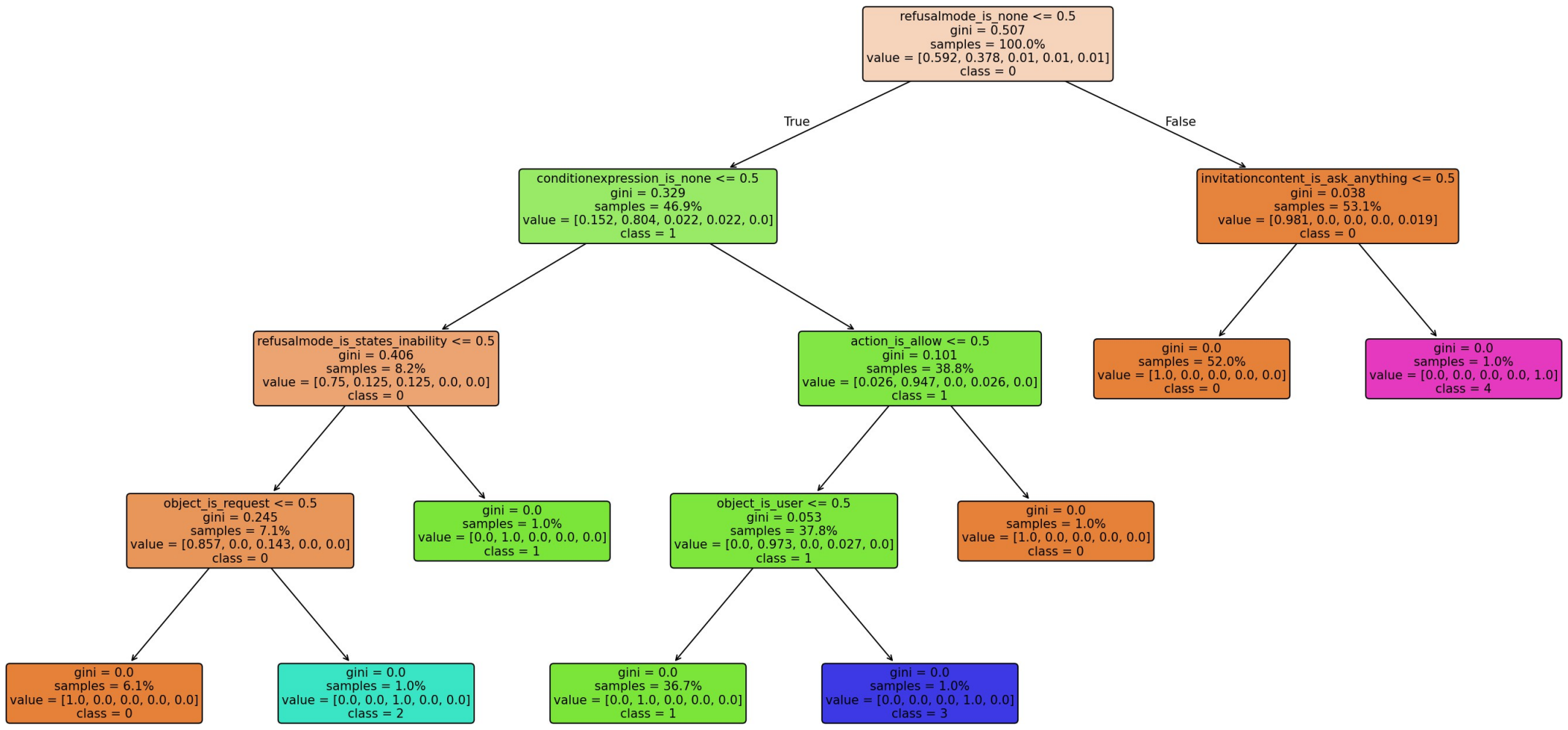

Figure 4: The decision tree computed for the AccessControl example.

### A.6 Corrective Statement Derivation

An LLM is prompted with the node's current system prompt, the elicited feature classes and their importance scores, the decision tree rules, and the description of the passing cluster. Nodes whose tree does not exceed the cv-F1 gate ($\geq$ 0.6) are skipped. The model returns corrective statements to be injected into the node's system prompt.

```
You are an expert system prompt engineer analyzing agent execution patterns.

You have access to:
1. The current system prompt being evaluated
2. Semantic features extracted from agent traces (Subject, Verb, Object, ...)
3. Decision tree analysis showing which features differentiate successful
   from failed traces
4. Feature importance scores

Your task is to recommend system prompt improvements that would guide agents
toward success patterns.

## CURRENT SYSTEM PROMPT: ...
## DISCOVERED SEMANTIC FEATURES: ...
## DECISION TREE RULES: ...
## FEATURE IMPORTANCE: ...
## SUCCESS CLUSTER: ...
```

Listing 3: Corrective-statement generation prompt (abridged).

For the AccessControl example, the corrective statements resolve the conditional ambiguity in the original prompt:

- Only allow a user if they are NOT both unauthorized AND untrusted.
- If either condition (unauthorized or untrusted) is true, refuse the request.

## B PIPELINE IMPLEMENTATION DETAILS

Table 5 lists the pipeline configuration details for the reported experiments. As noted in §3.2, the SFA pipeline supports two regimes: labelled and unlabelled; the latter requires user judgement to identify the desirable cluster, as in the illustrated example above. The decision tree's classification target is a binary run outcome corresponding to the labels. The pipeline resolves it in priority order: stored labels, the benchmark's pass/fail rating on the root span, an operator-supplied CSV, and finally, if none of these exist, spectral clustering of the answer-node outputs followed by a single user judgement identifying the desirable cluster. In the experiments reported in this paper the clustering of the answer-node outputs is skipped, as the labels come from the benchmark rating.

Embeddings are used for a different purpose: discretising elicited feature values. Each feature class is instantiated by free text across runs; those instantiations are embedded and grouped so that each class becomes a small set of discrete values, producing the binary feature matrix.

The cv-F1 threshold of 0.6 is a conservative choice, not a tuned one. A no-skill classifier sits near the macro-F1 floor, so this value separates nodes whose features carry real discriminative signal from those at chance. The threshold was not swept; a sensitivity analysis is listed as future work in §7.

Every arm runs the same agent model (`claude-haiku-4-5`). SFA's corrective statement derivation model is `claude-sonnet-4-6`, which is also the reflection model used by GEPA and `control` and the instruction proposer for MIPROv2. No arm is given a stronger model than any other for either execution or optimisation.

Table 5: Pipeline configuration for the reported experiments.

| **Stage** | **Setting** | **Value** |
|---|---|---|
| Run labels | source (all reported runs)<br>fallback (unlabelled use) | benchmark pass/fail rating on the root span<br>spectral clustering of answer-node outputs + one judgement |
| Feature-value encoding | embedding model<br>grouping<br>matrix | `nli-distilroberta-base-v2`<br>$k$-means, $k$ by silhouette score<br>binary: one column per feature value, 1 if present |
| Feature ranking | classifier<br>target<br>gate | CART decision tree, Gini impurity<br>binary run outcome<br>cross-validated macro-F1 ≥ 0.6 |
| Models | distillation, feature elicitation<br>corrective statements, rule assembly<br>agent under study | `claude-haiku-4-5`<br>`claude-sonnet-4-6`<br>`claude-haiku-4-5` |
| Sampling | instances per label for elicitation<br>trace selection | 20<br>balanced sample, retaining all-fail tasks |

A schematic illustration of the pipeline end-to-end on one real node from the benchmark study (HoVer, depth 2, summarize1) is shown in Figure 5. Two traces (one failing, one passing) are distilled, decomposed into feature values and separated by the tree; the decisive signal becomes the injected rule. Every string shown is taken from the recorded run.

## C OUTCOME FAILURE TAXONOMY

As a complementary qualitative investigation aimed at corroborating the attribution of SFA as an aid to resolving prompt ambiguity, the failure taxonomy presented here was derived bottom-up from the pipeline's own output. First, each of the 1200 training traces was distilled individually by an LLM into a single natural-language story together with its pass/fail label. Second, those per-trace stories were aggregated across runs: the pipeline elicited semantic feature classes from the cross-trace vocabulary and encoded each trace as a boolean feature vector. Third, the taxonomy rows were

---

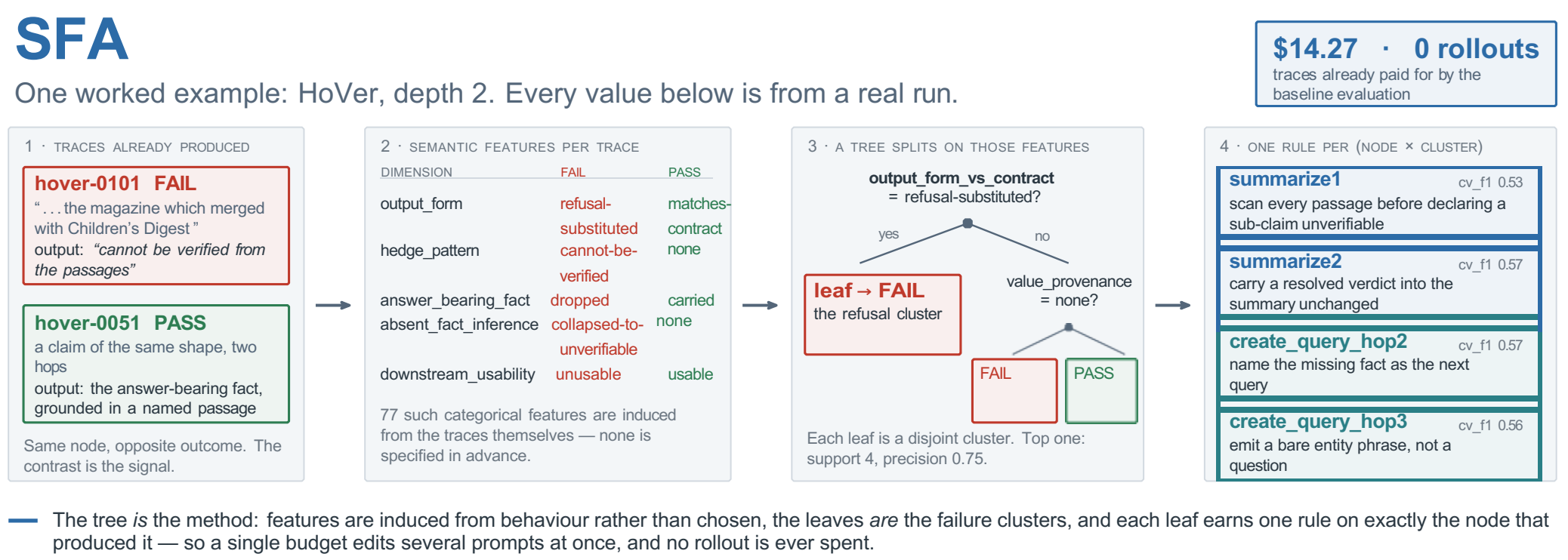

Figure 5: SFA end-to-end on one real node from the benchmark study (HoVer, depth 2, summarize1).

named from that feature vocabulary, bottom-up from the elicited classes. Fourth, a deterministic detector was written for each class to recompute both share columns from the stored artifacts.

The two share columns measure how often each failure pattern appears in runs of each type. *Share on failure* is the proportion of failing runs in which the defect was detected. A high value indicates the class is a common cause of failure. *Share on success* is the proportion of passing runs in which the same pattern was also detected. A value comparable to the failure share indicates the pattern alone does not predict failure and that additional context is needed to explain it.

The dominating failures in the table (verification not performed, query construction, inter-node handoff, and output format) share a common root: the agent's system prompt left a behavioural expectation underdetermined. Each class reflects a prompt that did not specify what the agent should do in a particular situation. As argued in §6, these defects are a form of prompt ambiguity, and it is precisely that ambiguity that SFA targets: the divergence it causes in execution traces is the signal the pipeline harvests to derive corrective statements.

Table 6: Failure classes identified in the train corpora. Description gives the nature of the defect. Share on failure is the proportion of failing runs in which the class was observed; Share on success is the proportion of passing runs showing the same pattern.

| **Failure class** | **Description** | **Instance** | **Share on failure** | **Share on success** |
|---|---|---|---|---|
| Verification not performed | A required check is skipped; the node returns the input unchanged | IFBench: checker returns the draft unchanged | 68% | 55% |
| Query construction | The query issued to a downstream tool is malformed or under-constrained | HoVer: hop-3 emits a polar question, not keywords | 13% | 15% |
| | | HotpotQA: hop-2 query drops the entity anchor | 47% | 67% |
| Inter-node handoff | A key entity or result is lost when passing output to the next node | HotpotQA: summary drops the bridge entity | 32% | 55% |
| Cross-node redundancy | A node repeats work already done by a preceding node | HoVer: hop-3 re-retrieves hop-2's documents | 80% | 92% |
| Output format | The answer is correct but surrounded by explanatory text instead of being returned directly | HotpotQA: gold value wrapped in prose | 45% | 0% |
| Over-refusal | The agent declines to answer when a correct answer exists | HotpotQA: "cannot be determined" with gold present | 2% | 0% |
| Tool-failure handling | The agent has no recovery strategy when a tool returns an error | GAIA: web_fetch returns PDF/403/JS | 61% | 30% |